\def\year{2021}\relax
\documentclass[letterpaper]{article} 
\usepackage{aaai21}  
\usepackage{times}  
\usepackage{helvet} 
\usepackage{courier}  
\usepackage[hyphens]{url}  
\usepackage{graphicx} 
\usepackage{natbib}  
\usepackage{caption} 
\usepackage{multirow}
\usepackage{multicol}
\usepackage{algorithm}
\usepackage{algorithmic}
\newcommand{\ie}{{\it i.e.}}

\usepackage{amsmath}

\title{Deep Epidemiological Modeling by Black-box Knowledge Distillation: \\An Accurate Deep Learning Model for COVID-19 }

\author {
    Dongdong Wang,\textsuperscript{\rm 1}
    Shunpu Zhang, \textsuperscript{\rm 2}
    Liqiang Wang \textsuperscript{\rm 1} \\
}
\affiliations {
    \textsuperscript{\rm 1} Department of Computer Science, University of Central Florida \\
    \textsuperscript{\rm 2} Department of Statistics and Data Science, University of Central Florida\\
    daniel.wang@knights.ucf.edu, \{Shunpu.Zhang, Liqiang.Wang\}@ucf.edu
}

\begin{document}

\maketitle

\begin{abstract}

An accurate and efficient forecasting system is imperative to the prevention of emerging infectious diseases such as COVID-19 in public health. This system requires accurate transient modeling, lower computation cost, and fewer observation data. To tackle these three challenges, we propose a novel deep learning approach using black-box knowledge distillation for both accurate and efficient transmission dynamics prediction in a practical manner. First, we leverage mixture models to develop an accurate, comprehensive, yet impractical simulation system. Next, we use simulated observation sequences to query the simulation system to retrieve simulated projection sequences as knowledge. Then, with the obtained query data, sequence mixup is proposed to improve query efficiency, increase knowledge diversity, and boost distillation model accuracy. Finally, we train a student deep neural network with the retrieved and mixed observation-projection sequences for practical use.  The case study on COVID-19 justifies that our approach accurately projects infections with much lower computation cost when observation data are limited.
\end{abstract}


\section{Introduction}

The spread of infectious diseases is a serious threat to public health and may cause million deaths every year. To effectively battle against infectious diseases, accurate modeling on their transmission patterns is critical. This issue becomes more pressing when the infectious disease, like COVID-19, is unprecedented, transmission dynamics is complex, and observation data are limited. Due to data limitation, we need to solve this problem with the help of conventional physics-based epidemiological models. However, it is still difficult to accurately describe complex dynamics with a single model.





Mixture models are widely used to accurately solve complex transient modeling problems. They can refine temporal scale into several states with different onsets, model these states separately, and then mix modeling results to represent complex dynamics. Although this refinement on temporal scale more accurately depicts the variation in a physical system, the difficulty of calibrating a mixture model and computational complexity can exponentially increase since it can result in very large parameter space, \ie, curse of dimensionality. When prior knowledge about an infectious disease, such as COVID-19, is limited, exhaustive search in such large space is inevitable for accurate model calibration, which can easily render a mixture model impractical. In reality, some modelers propose some assumptions to truncate search space with coarse grid and trade for efficiency and feasibility, but it can cause large uncertainty and model degradation. 

To address this problem, we formulate a new approach with black-box knowledge distillation. This approach is developed based on three-fold objectives, including higher prediction accuracy, lower modeling cost, and higher data efficiency. To achieve higher prediction accuracy, we first leverage mixture models to create a comprehensive, accurate, but probably impractical epidemic simulation system. This system is viewed as a black-box teacher model which contains sophisticated modeling knowledge. To reduce modeling cost and make this system feasible, we employ knowledge distillation to transfer the accurate modeling knowledge from this impractical black-box teacher model to a deep neural network for practical use. To realize this knowledge transfer, we collect a set of simulated observation sequences to query the teacher model and acquire their corresponding simulated projection sequences as knowledge. Particularly, for improvement in model performance with limited data, we propose sequence mixup to augment data pool, thus reducing model queries, increasing sequence diversity, and boosting modeling accuracy. With all retrieved and mixed observation-projection sequence pairs, we train a student deep neural network for infection prediction. This student network can perform prediction as accurately as teacher model, but save lots of computation cost, and require fewer observation data.



To the best of our knowledge, we are the first to propose a black-box knowledge distillation based framework to solve epidemiological modeling by leveraging mixture models. Besides this novelty, our work also includes the following contributions: (1) the distilled student deep neural network enables accurate model calibration and projection automatically. (2) Sequence mixup is proposed to reduce teacher model queries for higher efficiency, improve the coverage of obtained data for better accuracy, and further enhance knowledge transfer with fewer observation data. (3) We justify our approach by solving COVID-19 infection projection and it performs on par or even better than some state-of-the-art methods, like CDC Ensemble, with adequate accuracy over the evaluation period. (4) Our approach provides a general solution to render impractical physics-based models feasible. 

\section{Related Work}

\subsection{Epidemiological Modeling}

Epidemiological modeling has been extensively studied for decades. It is focused on how to accurately quantify infectious disease transmission dynamics. The proposed methods can be classified into two main categories, classical physics-based modeling and data-driven approach. For physics-based modeling, compartmental modeling, like SEIR \cite{kermack1927contribution}, is well justified for practical projection. Different from physics-based modeling, thanks to the improvement on data collection, data-driven approaches have been developed based upon statistical modeling on real observation data and widely used for transmission dynamics projection, such as ARIMA\cite{benjamin2003generalized} and ARGO\cite{yang2015accurate,yang2017using}. With rapid advances in artificial intelligence, deep learning based modeling as an alternative is proposed to solve infection projection, especially for emergency pandemic like COVID-19 \cite{wu2020nowcasting,hu2020artificial,yang2020modified,fong2020composite}. However, these data-driven approaches can suffer from observation data limitation. Recently, a hybrid approach named DEFSI \cite{wang2019defsi} adopts compartmental modeling to alleviate data limitation problem in deep neural network training. 

\subsection{Knowledge Distillation}

Knowledge distillation \cite{hinton2015distilling} is widely used to solve deep neural network compression problem. Conventional distillation process is carried out by training a smaller neural network called student model with class probability, which is referred to as ``dark knowledge", to retain the performance of original cumbersome ensemble of models called teacher model. This approach can effectively reduce model size, which makes complex models feasible for real-world applications. Many complex applications in computer vision or natural language processing have justified its merits for model size reduction. For example, DistilBERT \cite{sanh2019distilbert} successfully reduces the size of original BERT model by $40\%$ with maintaining accuracy; TinyBERT \cite{jiao2019tinybert} leverages knowledge distillation to design a framework for the reduction of transformer-based language model, which leads to the models with lower time and space complexity, thus facilitating its application; relational knowledge distillation \cite{park2019relational} further optimizes distillation process and enables more productive student model, which can even outperform teacher model. However, this effective approach has not been applied to solve complex epidemiological modeling, especially the infeasibility of mixture epidemiological models.

\subsection{Mixup}
Mixup is a simple yet effective approach to augment training data and improve model performance \cite{zhang2017mixup}. This method is proposed to improve the generalization of deep neural network by enhancing coverage of data distribution, especially when training data are limited. The main idea is to incorporate convex combination into data synthesis, which involves mixing features and mixing labels. It has been widely used to address computer vision and natural language processing problems, like Between-Class learning in speech recognition \cite{tokozume2017learning} and image classification\cite{tokozume2018between}, AutoAugment with learning strategy augmentation for classification \cite{cubuk2018autoaugment}, and wordMixup or senMixup with embedding mixup for sentence classification \cite{guo2019augmenting}. More studies explore its potential for data-efficient learning, such as active mixup \cite{wang2020neural} and ranking distillation in \cite{laskar2020data}. However, there is no work using mixup to enhance epidemiological modeling efficacy and efficiency.

\begin{figure}[t]
    \centering
    \includegraphics[width=0.44\textwidth]{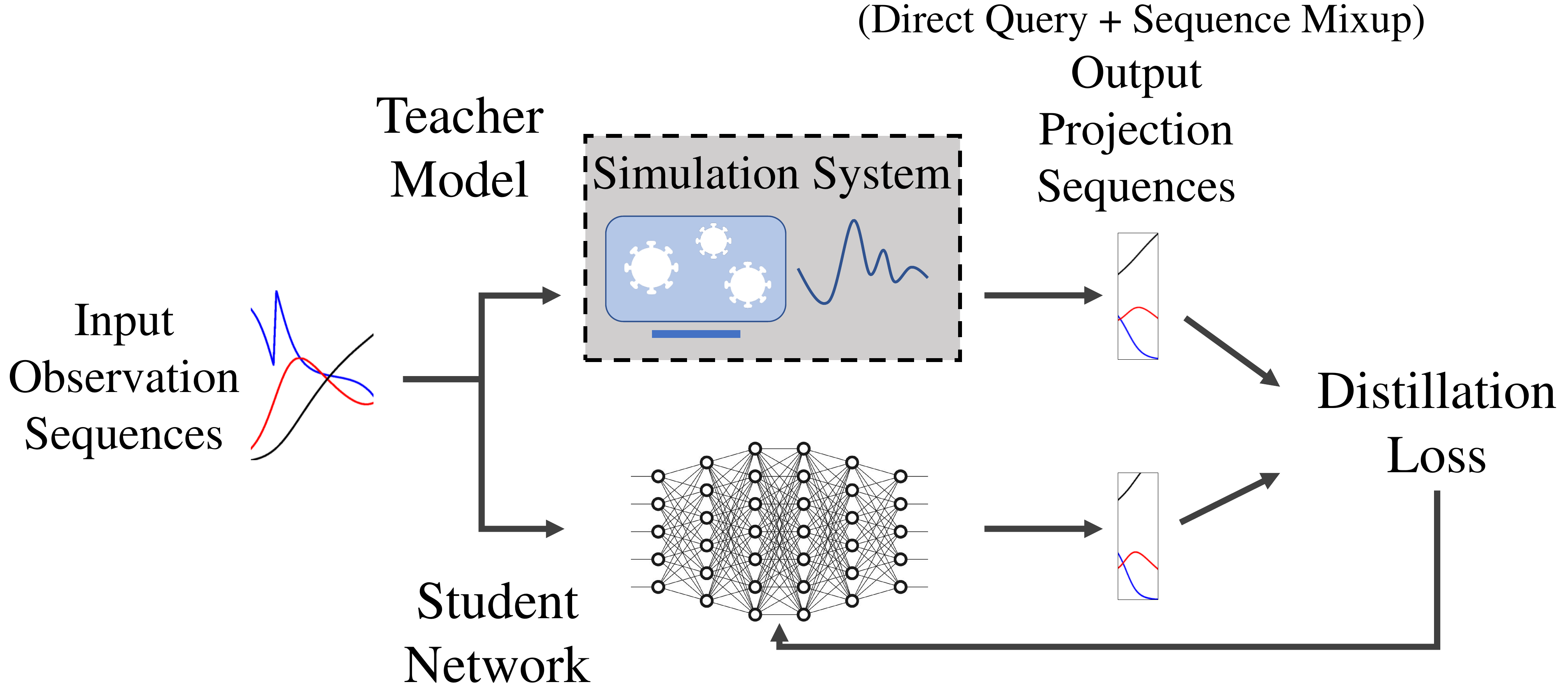}
    \caption{Modeling with black-box knowledge distillation. Teacher model is an accurate but significantly complex comprehensive simulation system. Both observation and projection sequences are simulated results. Model query is optimized by sequence mixup.}
    \label{fig:framework}
\end{figure}

\section{Methodology}

Figure \ref{fig:framework} shows an overview of our approach on epidemiological modeling by black-box knowledge distillation. We leverage mixture models to build a comprehensive simulation system with accurate modeling knowledge yet significantly high complexity. Then, we use simulated observation sequences to query this system to retrieve simulated projection sequences as knowledge. To improve query efficiency and enhance knowledge transfer, sequence mixup is designed to further efficiently augment data pool. With retrieved and mixed observation-projection sequence pairs, a deep neural network is trained to retain the modeling accuracy of the original impractical simulation system and prepared for practical use.

\subsection{Developing a Teacher Model}

Many approaches can be used to create mixture models and build a comprehensive simulation system $\mathcal{M}$. To ensure reliability, we select a widely accepted compartmental model of SEIR as the modeling approach. In SEIR, people in the modeled society, aka host society, must be in one of the four health states, \ie, susceptible, exposed, infectious, and recovered. The state transition starts from ``susceptible", and then moves to ``exposed", then to ``infectious", and finally reaches ``recovered" state. Thus, the model is constrained with the boundary condition of $N$ = $S$ + $E$ + $I$ + $R$, where $S$, $E$, $I$, and $R$ denote susceptible, exposed, infected, and recovered population, respectively, and $N$ represents the population of the entire host society. 

For accurate depiction of transient transmission dynamics, we employ linear mixture model \cite{brauer2017mathematical} to represent the heterogeneity of host society \cite{bansal2007individual}. The host society $N$ is divided into several component host communities $N_i$ with the linear combination in Equation \ref{eq:sep}, and modeling results from these communities will be mixed to represent the dynamics of entire host society $N$. The division of host society is based on heuristics, which depends on modeling resolution. 
\begin{equation}
\label{eq:sep}
N  =  \sum_{i=0}^n N_i =  \sum_{i=0}^n (S_i  + E_i + I_i + R_i) \\
\end{equation}

Within each community $N_i$, transmission dynamics can be described by an ordinary differential equation (ODE) system, as shown in Equation \ref{eq:SEIR}, across all compartments.
\begin{equation}\label{eq:SEIR}
\begin{split}
\frac{dS_i}{dt} & =  \alpha N_i - \beta S_i^t I_i^t - \mu N_i S_i^t  \\
\frac{dE_i}{dt} & =  \beta S_i^t I_i^t - (\sigma + \mu) E_i^t \\
\frac{dI_i}{dt} & =  \sigma E_i^t - (\gamma + \mu) I_i^t  \\
\frac{dR_i}{dt} & =  \gamma I_i^t - \mu R_i^t 
\end{split}
\end{equation}
where $S_i^t$, $E_i^t$, $I_i^t$, and $R_i^t$ denote susceptible, exposed, infected, and recovered population, respectively, at time $t$. $\beta$, $\sigma$, and $\gamma$ denote infectious, latent, and recovery rate over the entire incidence, respectively. $\alpha$ and $\mu$ are referred to as natural birth and death rates during this period, respectively, which are assumed to be zero in this study. 

SEIR modeling is a typical boundary value problem \cite{farlow1993partial}, the solution of which relies on boundary condition (BC), initial condition (IC), and ODEs. In this study, for each component host community, constant BC is assigned by the total population $N_i$ due to no vital dynamics, IC is determined by the compartment state information $\{S_i^0, E_i^0, I_i^0, R_i^0\}$ at time step $t=0$, and ODEs are specified by the dynamics coefficients $\{\beta, \sigma, \gamma\}$. Conventional numerical modeling requires model calibration, , which adjusts parameters to obtain agreement between real observation data and modeled results, using grid search for an optimal combination of BC, IC, and ODEs ($\{$BC, IC, ODEs$\}$) within constraints in search space. If the search space for $\{$BC, IC, ODEs$\}$ is larger and fine-grained, the calibration results are better fit to the real observation data and simulated projected results are more reliable. Therefore, we construct a comprehensive simulation system with an ensemble of simulation scenarios from large and fine search space, which enables accurate model calibration and projection. 

However, the complexity of this simulation ensemble system is very time-consuming for grid search due to curse of dimensionality. For example, suppose we have just $2$ options for BC, IC, and ODEs (the real problems require much more). For each component host community, there are $8$ simulation scenarios. However, if we have 10 component communities, the ensemble for the entire society $N$ will reach $8^{10}$ simulation scenarios. It is infeasible to find an optimal solution with random grid search. Therefore, we conduct knowledge distillation to distill this ensemble simulation system into a deep neural network for practical use.

\subsection{Querying the Teacher Model}

Conventional knowledge distillation is carried out by querying the teacher model to obtain prediction probabilities that are referred to as ``knowledge". In our problem, the ``knowledge" are simulated projection sequences from the simulation system since they contain the features of modeling process. To facilitate acquiring such kind of modeling ``knowledge", we conduct model querying as follows.
First, we prepare a simulated observation sequence over the calibration period with a $\{$BC, IC, ODEs$\}$ for each host community. Each $\{$BC, IC, ODEs$\}$ is used as a ``key" to query teacher model. Then, the teacher model will use the ``key" to return a query answer with a simulated sequence over the calibration and projection period, \ie, a projection sequence. With more queries, more projection sequences are obtained and more accurate modeling knowledge is acquired.


\begin{figure*}[t]
    \centering
        \includegraphics[width=0.98\textwidth]{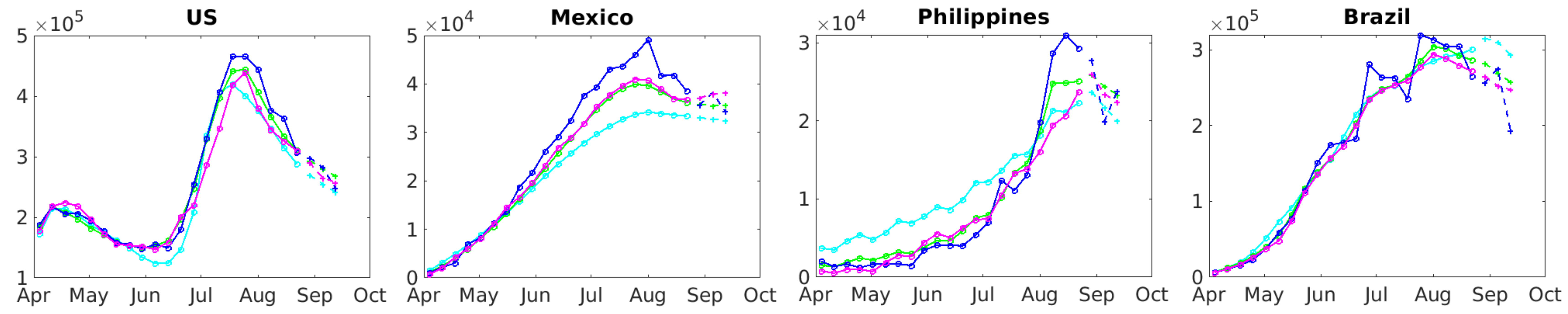}
        \includegraphics[width=0.50\textwidth]{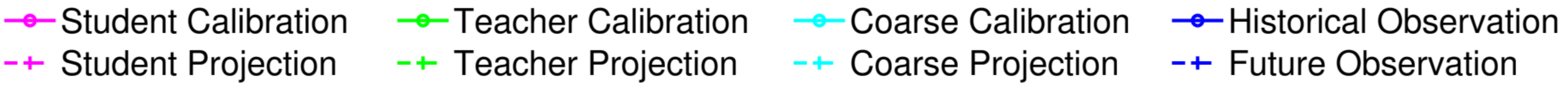}
     \caption{Weekly new infection cases over the calibration (04/06-08/23) and projection (08/24-09/13) periods by teacher model, student network, and coarse search. }
     \label{fig:results}
\end{figure*}

\subsection{Sequence Mixup}

To ensure adequate knowledge, distillation usually requires lots of training data from many model queries. However, too many queries can be time-consuming, and more importantly, the simulated observation sequences are still too limited to acquire diverse knowledge. For improvement in distillation efficacy and data diversity, we employ sequence mixup to reduce the number of queries and enlarge knowledge coverage. 

\begin{equation} \label{eq:seq_mix}
\begin{split}
\hat{x} & = \omega_1 x_1 + \omega_2 x_2 + ... + \omega_n x_n
\\
\hat{y} & = \omega_1 y_1 + \omega_2 y_2 + ... + \omega_n y_n
\end{split}
\end{equation}

Our sequence mixup is developed with convex combinations of multiple observation sequences $x_i$ and projection sequences $y_i$ with mix rates $\omega_i$, where $\Sigma\omega_i = 1$. Equation \ref{eq:seq_mix} presents this mixup process which mixes observation sequences $x$ and projection sequences $y$ in the same manner.


\begin{equation} \label{eq:proof}
\begin{split}
S^{t+1} & = S^t + \frac{dS^t}{dt} = \sum_{i=1}^{n} \omega_i S_i^t + \frac{d\sum_{i=1}^{n} \omega_i S_i^t}{dt} \\
& = \sum_{i=1}^{n} \omega_i S_i^t + \sum_{i=1}^{n} \frac{d \omega_i S_i^t}{dt} = \sum_{i=1}^{n} \omega_i S_i^t + \frac{d \omega_i S_i^t}{dt} \\
& = \sum_{i=1}^{n} \omega_i S_i^{t+1} 
\end{split}
\end{equation}

 The mixup projection sequence $\hat{y}$ in Equation \ref{eq:seq_mix} uses the same coefficients $\omega_{1} , \omega_{2} , ... , \omega_{n}$ as in $\hat{x}$ and it can be briefly proved as follows. Suppose $\hat{x}$ denotes $S^t = \sum_{i=1}^{n} \omega_i S_{i}^{t}$ at the current observation time and $\hat{y}$ denotes $S^{t+1}$ at the next projection time. Given the linearity of differentiation, this mixup process $S^{t+1} = \sum_{i=1}^{n} \omega_i S_i^{t+1}$ is justified in Equation \ref{eq:proof}. Similar proof can be completed for $E$, $I$, and $R$.

These mixed sequences as an alternative to query knowledge efficiently augment training data and enhance the knowledge transfer from teacher model. Thus, all retrieved and mixed sequences construct a training set $(X, Y)$.

\subsection{Training a Student Deep Neural Network}

With the acquired observation-projection sequence pairs $(X, Y)$, a deep neural network is trained to distill the modeling knowledge within the comprehensive simulation system. The conventional distillation process is carried out by minimization on the distillation loss function $L_{dis} = D_1(y_n^{true}, S(x_n)) + D_2(T(x_n), S(x_n))$, where $T(x_n)$ is the output of data $x_n$ from teacher model $T$, $S(x_n)$ is the output of data $x_n$ from student network $S$, $D_1$ is the supervised loss for supervised learning with data label $y_n^{true}$, and $D_2$ is the imitation loss for model output imitation. In our problem, there is no knowledge about the true label $y_n^{true}$ for $x_n$, and thus, the distillation loss is modified to the imitation loss only, as shown in Equation \ref{eq:dis_loss}. We select mean squared error loss as distillation loss function.
\begin{equation}
    L_{dis} = D_2(T(x_n), S(x_n))
    \label{eq:dis_loss}
\end{equation}

The proposed black-box knowledge distillation is a general approach that can be applied to different student networks. In the problem of COVID-19, we use multilayer perceptron (MLP) which is detailed in the case study.

\begin{table*}[t]
\centering
\begin{tabular}{c|c|cccc|cccc}
\cline{1-10}
\multirow{2}{*}{Metric}  & \multirow{2}{*}{Model} & \multicolumn{4}{c|}{Calibration} & \multicolumn{4}{c}{Projection}\\\cline{3-10}
 &  & US & Mexico & Philippines & Brazil &  US & Mexico & Philippines & Brazil  \\ \hline

\multirow{3}{*}{MAPE} & Teacher & \textbf{0.0363} & 0.1217 & \textbf{0.3197} & 0.0879&  \textbf{0.0352} & \textbf{0.0369} & 0.1030 & 0.1522 \\
 & Student &  0.0695 & \textbf{0.1164} & 0.3472 & \textbf{0.0792} &  0.0433 & 0.0527 & \textbf{0.0984} & \textbf{0.1331} \\
 & Coarse & 0.0843 & 0.2269 & 1.3159 & 0.1438 &  0.0727 & 0.0910 & 0.1314 & 0.2923 \\\hline
  & Teacher &  \textbf{0.669} & 0.183 & \textbf{0.101} & \textbf{0.790} &  \textbf{0.209} & \textbf{0.028} & 0.048 & 0.703 \\
 RMSE & Student & 1.321 & \textbf{0.163} & 0.163 & 0.857 &  0.218 & 0.041 & \textbf{0.041} & \textbf{0.593} \\
($10^5$)& Coarse & 1.426 & 0.333 & 0.229 & 0.985 &  0.399 & 0.063 & 0.059 & 1.215 \\
\cline{1-10}
\end{tabular}
\caption{Error assessment of model calibration (04/06 - 08/23) and projection (08/24 - 09/13).}
\label{tab:accuracy}
\end{table*}

\subsection{Overall Algorithm}

Algorithm \ref{algo:KD} presents the overall procedure of our proposed black-box knowledge distillation based epidemiological modeling. Beginning with a modeling approach, a comprehensive epidemic simulation system is built as a teacher model $\mathcal{M}^T$. We then pick a few simulated observation sequences $x$ to query the teacher model and retrieve their simulated projection sequences $y$. With obtained sequences $(x, y)$, we construct a large observation-projection pool $(X,Y)$ using sequence mixup. Finally, we train a student deep neural network $\mathcal{M}^S$ with $(X,Y)$.
\begin{algorithm}[h]
\caption{Epidemiological Modeling with Black-box Knowledge Distillation }
\hspace*{0.02in}
\label{algo:KD}

{\bf INPUT:} A modeling approach $F$ such as mixture SEIR.

{\bf INPUT:} A set of observation sequences $X_{obs} = \{x_i\}^n_{i=1}$.

{\bf INPUT:} Hyper-parameters (mixup rate, learning rate etc.)

{\textbf{OUTPUT:}}  A student deep neural network $\mathcal{M}^S$ 

\begin{algorithmic}[1] 
\STATE{Develop a comprehensive simulation system $\mathcal{M}^T$ based upon $F$ with a set of conditions $\{$BC, IC, ODEs$\}$}s
\STATE{With all observation sequences in $X_{obs}$, query simulation system $\mathcal{M}^T$, retrieve projection sequences $Y_{query} =\{y_i\}^n_{i=1}$, and form an observation-projection pool $(X_{obs},Y_{query})$.}
\STATE{Construct a mixed sequence pool $(X_{mix},Y_{mix}) = \{  (\hat{x}, \hat{y}) : (\hat{x}, \hat{y}) \in (\sum_{i=1}^n \omega_i x_i , \sum_{i=1}^n \omega_i y_i) \}$  with query results $(X_{obs}, Y_{query})$}, where $\omega$ is heuristically chosen.

\STATE{Train a student deep neural network $\mathcal{M}^S$ with $(X,Y) = (X_{obs},Y_{query}) \cup (X_{mix},Y_{mix}) $ to minimize distillation loss $L_{dis}$.}\\

\end{algorithmic}
\end{algorithm}

\section{COVID-19 Case Study}

\subsection{Experiment Setting} 

\subsubsection{Data.} 
We evaluate our approach on the open COVID-19 datasets provided by Johns Hopkins University \cite{dong2020interactive}. In this dataset, our experiments are focused on daily infection case increase. With these reported data, we derive active infection cases based on 7-day transmission duration \cite{thevarajan2020breadth}, as the data do not explicitly report the number of recovered patients. The observation period starts from 04/06/2020 to 08/23/2020 and the evaluation period is from 08/24/2020 to 09/13/2020. 

\subsubsection{Black-box Teacher Model.}
A black-box teacher model is built with aforementioned mixture SEIR. The mixture model consists of 10 compartment host communities. Each compartment host community is simulated with 10 choices for $N_i$ to specify constant BC, 2 choices for $\{S_i^0, E_i^0, I_i^0, R_i^0\}$ to specify IC, and 20 choices for each coefficient in $\{\beta, \sigma, \gamma\}$ to specify ODEs. Such choices of parameters are based on heuristics. Most studies on COVID-19 using SEIR model give a wide range of parameter choices \cite{liu2020reproductive}. We refine them to more reliable ranges. With the refined parameter choices, this simulation system contains $160000^{10}$ scenarios for the entire society $N$, which is impractical. To facilitate distillation assessment, we conduct random sampling to reduce it to $10^{7}$ scenarios as an approximate version of teacher model to the simulation system for comparative study. The teacher model generates a simulated projection sequence by minimizing the mean squared error between real observation and the simulation over the calibration period, which is similar to exhaustive search. 

\subsubsection{Query Sequences and Mixup.}
We randomly pick 1000 $\{$BC, IC, ODEs$\}$s to prepare simulated observation sequences which are used to query teacher system. Note that, compared to the size of the ensemble, this number is so limited that we acquire little knowledge about simulation system with selected sequences, which still follows black-box teacher model setting. Given 1000 query results, we construct a large pool with 100K sequences by sequence mixup, where $\omega$ is set heuristically.

\subsubsection{Student Deep Neural Network Training.}
\label{subsec:student}
Our student network architecture is an MLP which has 3 hidden layers with 80 neurons each. The batch size is 128 and learning rate is set to 0.1. Adam optimizer is chosen. Weight decay is specified to 1e-5. The total epoch is set to 300 and learning rate is reduced by 90$\%$ after every 100 epochs. We select 1K sequences from the constructed sample pool as a training set for efficient training.

\subsubsection{Studied Cases.}
We implement our black-box distillation framework to distill comprehensive infection modeling system for US, Mexico, Philippines, and Brazil. The infection patterns of these countries are representative of complex dynamics which involves multiple peaks and complicates model calibration. To achieve an adequate teacher model on each studied country, we heuristically specify the search space boundaries for $\{$BC, IC, ODEs$\}$s with the information of national population, reported positive cases on March 30th (a week before April 6th), and outbreak severity for each country.

\subsubsection{Evaluation Metric.}
 We evaluate infection case modeling performance on both accuracy and efficiency. For accuracy, model calibration and projection are assessed. The performance is quantified by mean absolute percentage error, MAPE $ = \frac{1}{n} \sum^n_{i=1}|\frac{y^o_i - y^m_i}{y^o_i}| \label{eq:re_formula}$, and root mean square error,  RMSE $= \sqrt{\frac{1}{n} \sum^n_{i=1}(y^o_i - y^m_i)^2}$, where $y^o$ is the real observation sequence, $y^m$ is the modeled sequence, and $n$ is the total number of sequences. MAPE and RMSE are two widely adopted metrics to evaluate regression models. While lower MAPE suggests that the general trend is better captured, higher error can occur at larger observation data. RMSE is a better indicator for large values since it offers higher penalty for these errors. Therefore, we use both metrics for accuracy evaluation. 
 
 As to computation efficiency, we evaluate model complexity with required simulation scenarios and total time cost for each projection query. For student network, the network training cost is included in each query process although network retraining is not always necessary. 


\begin{table*}[h]
\centering
\begin{tabular}{c|cccccccc}
\hline
Period & \multicolumn {8}{c}{Model}\\ \cline{2-9}
(from 08/23)   &  CDC Ensemble & UM   & DDS & UVA & UCLA & JHU & Columbia & Ours (Student) \\ \hline
1 week ahead & 0.0608 & 0.3866 & 0.0417 & 0.0698 & 0.0367 & 0.0737 & 0.0456 & \textbf{0.0301}\\
2 week ahead & 0.1108 & 0.0386 & \textbf{0.0228} & 0.0772 & 0.0889 & 0.1165 & 0.0250 & 0.0623\\
3 week ahead & 0.0581 & 0.0549 & 0.0819 & 0.2724 & \textbf{0.0077} & 0.2572 &0.2083 & 0.0398\\

\cline{1-9}
\end{tabular}
\caption{MAPE comparison of state-of-the-art models and our method on US weekly infection case increase projection between 08/24 and 09/13. The results of other models are collected from CDC, which are reported by COVID-19 Forecast Hub.}
\label{tab:benchmark}
\end{table*}



\subsubsection{Competing Methods.}

First, we compare our approach with the approximate teacher model and coarse search to examine accuracy and efficiency. Coarse search is developed upon coarse grid search space for mixture models. We reduce the number of compartment communities to 5, the options for BC to 5, and the choices for each ODE coefficient to 10, which could be taken as a reduced teacher model, but still with the complexity of $10000^{5}$. Similar to teacher model, for practical performance evaluation, we reduce it to $10^5$ scenarios with random sampling, which ensures its similar data complexity to student network. In the following sections, approximate teacher model and coarse grid search are referred to as teacher model and coarse search, respectively. Next, we compare our student network with 7 state-of-the-art forecasting models reported from CDC \cite{bracher2020evaluating}. These models are developed with machine learning based methods, like UM and UCLA-SuEIR, statistical methods, like DDS, physics-based model, like JHU-IDD and Columbia, and ensemble approaches, like UVA and CDC Ensemble \cite{ray2020ensemble}.


\subsection{Results}

\subsubsection{Accuracy.} Our calibration and projection results are reported with weekly increase cases in Figure \ref{fig:results}. Student network is comparable to teacher model and significantly outperforms coarse search. These performance differences are quantified with MAPE and RMSE in Table \ref{tab:accuracy}. It is shown that, compared to the teacher model, student network achieves similarly low or even lower MAPE and RMSE, over the calibration or projection periods. This observation results from the approximation of teacher model and sequence mixup for student network training. Coarse search yields highest errors due to limited search space. 

We compare our student network with 7 state-of-the-art models in Table \ref{tab:benchmark}, which are based on the reported data from CDC \cite{bracher2020evaluating}. Our model consistently outperforms CDC Ensemble, which incorporates all reported state-of-the-art models, with 30$\%-$50$\%$ MAPE reduction over this period. In particular, our model yields more accurate 1 week ahead prediction and more consistent performance over three weeks compared to other models.

\subsubsection{Efficiency.} From Table \ref{tab:complexity}, student network saves both simulations and time cost by orders of magnitude. Student network and coarse search are on par in total time cost, while the network training takes approximately 300 CPU seconds in our study. This performance gain results from the optimization with sequence mixup and lightweight network design. It justifies that our approach significantly improves modeling efficiency and can facilitate the application of complex and cumbersome epidemiological models.

\subsubsection{Significance of Mixup.}

Sequence mixup, as an efficient method for data augmentation, is very important to enhance knowledge transfer in our approach. Compared to coarse search and teacher model, our student network can learn more scenarios out of search space due to sequence mixup, and this knowledge can overcome the limit from search space, thus even improving calibration and projection accuracy. To justify its importance, we conduct experiments with 100K, 50K, and 25K mixed sequences from 1000 retrieved observation-projection sequences and evaluate their performance difference in calibration and projection for US. From Table \ref{tab:mixup_exp}, the reduction in mixed sequences causes model degradation. The degradation becomes worse in the projection period due to calibration error propagation. Thus, sequence mixup is critical to accurate projection.

\begin{table}[b]
\centering
\setlength{\tabcolsep}{3.5pt}
\begin{tabular}{ccccc}
\cline{1-5}
& Complete & Approximate & Student & Coarse \\
& Teacher &  Teacher & Network & Search \\\hline
Simulations & $160000^{10}$ & $10^7$ &  $10^3$ & $10^5$ \\
Time(s) & N/A & $\sim$3$\times10^4$  & $\sim400$ & $\sim$300 \\
\cline{1-5}
\end{tabular}
\caption{Model complexity measured by the required simulations and the CPU time cost for one projection query.}
\label{tab:complexity}
\end{table}

\begin{table}[h]
\centering
\begin{tabular}{llccc}
\cline{1-5}
      & Metric  &  100K & 50K & 25K \\ \hline
\multirow{2}{*}{Calibration} & MAPE & \textbf{0.0695} & 0.0987 & 0.1459  \\  & RMSE($10^5$)  & \textbf{1.321} & 1.831  & 2.910\\ \hline
\multirow{2}{*}{Projection} & MAPE & \textbf{0.0433} & 0.1861 & 0.2813 \\  & RMSE($10^5$) & \textbf{0.218} & 0.985  &  1.367  \\
\cline{1-5}
\end{tabular}
\caption{Calibration and projection errors from student network for US with 100K, 50K, and 25K mixed sequences.}
\label{tab:mixup_exp}
\end{table}

\subsubsection{Discussion.}

First, a comprehensive and accurate modeling system is critical in our framework. When this comprehensive teacher model is more complex and accurate, our student network can yield more accurate results. Next, student network can interpolate information in latent space which can resolve space discretization problem in grid search. The space of grid search is often too sparse to find an optimal solution. Therefore, dense search space is imperative, but its cost will exponentially increase. This can be alleviated by our proposed knowledge distillation. 
In addition, sequence mixup improves training data coverage and boosts model distillation, which helps student network even outperform teacher model. It implies that our proposed knowledge distillation scheme has potential to improve teacher model. Also, if a well-trained student network is obtained, the model could be reused many times, even when new data are included. In contrast, conventional random grid search, like teacher model or coarse search, has to be reset and query all entries again to retrieve projection solutions. This implies student network can save extra query cost.

\section{Conclusion}

We propose an innovative accurate modeling approach which leverages mixture models to ensure high accuracy and employs black-box knowledge distillation to reduce complexity and improve accuracy. It consists of teacher model development, model querying, sequence mixup, and student network training. The developed teacher model is a comprehensive simulation system which can accurately model challenged transient dynamics but is impractical. Then, we prepare simulated observation sequences to query this simulation system and retrieve simulated projection sequences as knowledge for distillation. In particular, to save number of queries and enhance knowledge transfer, sequence mixup is designed and effectively augments training data. With retrieved and mixed observation-projection sequences, a student deep neural network is trained as a distilled model for practical use. Our COVID-19 case study on US, Mexico, Philippines, and Brazil justifies that this approach brings in high accuracy but lower complexity. Also, our approach outperforms some state-of-the-art methods, like CDC Ensemble, over the studied period. In future, this work will be extended and applied to more epidemiological studies.

\section{Acknowledgements}
This project was support in part by NSF 1704309 and UCF COVID-19 Artificial Intelligence and Big Data Initiative.


\bibliography{main}

\begin{thebibliography}{28}
\providecommand{\natexlab}[1]{#1}
\providecommand{\url}[1]{\texttt{#1}}
\providecommand{\urlprefix}{URL }
\expandafter\ifx\csname urlstyle\endcsname\relax
  \providecommand{\doi}[1]{doi:\discretionary{}{}{}#1}\else
  \providecommand{\doi}{doi:\discretionary{}{}{}\begingroup
  \urlstyle{rm}\Url}\fi

\bibitem[{Bansal, Grenfell, and Meyers(2007)}]{bansal2007individual}
Bansal, S.; Grenfell, B.~T.; and Meyers, L.~A. 2007.
\newblock When individual behaviour matters: homogeneous and network models in
  epidemiology.
\newblock \emph{Journal of the Royal Society Interface} 4(16): 879--891.

\bibitem[{Benjamin, Rigby, and Stasinopoulos(2003)}]{benjamin2003generalized}
Benjamin, M.~A.; Rigby, R.~A.; and Stasinopoulos, D.~M. 2003.
\newblock Generalized autoregressive moving average models.
\newblock \emph{Journal of the American Statistical association} 98(461):
  214--223.

\bibitem[{Bracher et~al.(2020)Bracher, Ray, Gneiting, and
  Reich}]{bracher2020evaluating}
Bracher, J.; Ray, E.~L.; Gneiting, T.; and Reich, N.~G. 2020.
\newblock Evaluating epidemic forecasts in an interval format.
\newblock \emph{arXiv preprint arXiv:2005.12881} .

\bibitem[{Brauer(2017)}]{brauer2017mathematical}
Brauer, F. 2017.
\newblock Mathematical epidemiology: Past, present, and future.
\newblock \emph{Infectious Disease Modelling} 2(2): 113--127.

\bibitem[{Cubuk et~al.(2018)Cubuk, Zoph, Mane, Vasudevan, and
  Le}]{cubuk2018autoaugment}
Cubuk, E.~D.; Zoph, B.; Mane, D.; Vasudevan, V.; and Le, Q.~V. 2018.
\newblock Autoaugment: Learning augmentation policies from data.
\newblock \emph{arXiv preprint arXiv:1805.09501} .

\bibitem[{Dong, Du, and Gardner(2020)}]{dong2020interactive}
Dong, E.; Du, H.; and Gardner, L. 2020.
\newblock An interactive web-based dashboard to track COVID-19 in real time.
\newblock \emph{The Lancet infectious diseases} 20(5): 533--534.

\bibitem[{Farlow(1993)}]{farlow1993partial}
Farlow, S.~J. 1993.
\newblock \emph{Partial differential equations for scientists and engineers}.
\newblock Courier Corporation.

\bibitem[{Fong et~al.(2020)Fong, Li, Dey, Crespo, and
  Herrera-Viedma}]{fong2020composite}
Fong, S.~J.; Li, G.; Dey, N.; Crespo, R.~G.; and Herrera-Viedma, E. 2020.
\newblock Composite Monte Carlo decision making under high uncertainty of novel
  coronavirus epidemic using hybridized deep learning and fuzzy rule induction.
\newblock \emph{Applied Soft Computing} 106282.

\bibitem[{Guo, Mao, and Zhang(2019)}]{guo2019augmenting}
Guo, H.; Mao, Y.; and Zhang, R. 2019.
\newblock Augmenting data with mixup for sentence classification: An empirical
  study.
\newblock \emph{arXiv preprint arXiv:1905.08941} .

\bibitem[{Hinton, Vinyals, and Dean(2015)}]{hinton2015distilling}
Hinton, G.; Vinyals, O.; and Dean, J. 2015.
\newblock Distilling the knowledge in a neural network.
\newblock \emph{arXiv preprint arXiv:1503.02531} .

\bibitem[{Hu et~al.(2020)Hu, Ge, Jin, and Xiong}]{hu2020artificial}
Hu, Z.; Ge, Q.; Jin, L.; and Xiong, M. 2020.
\newblock Artificial intelligence forecasting of covid-19 in china.
\newblock \emph{arXiv preprint arXiv:2002.07112} .

\bibitem[{Jiao et~al.(2019)Jiao, Yin, Shang, Jiang, Chen, Li, Wang, and
  Liu}]{jiao2019tinybert}
Jiao, X.; Yin, Y.; Shang, L.; Jiang, X.; Chen, X.; Li, L.; Wang, F.; and Liu,
  Q. 2019.
\newblock Tinybert: Distilling bert for natural language understanding.
\newblock \emph{arXiv preprint arXiv:1909.10351} .

\bibitem[{Kermack and McKendrick(1927)}]{kermack1927contribution}
Kermack, W.~O.; and McKendrick, A.~G. 1927.
\newblock A contribution to the mathematical theory of epidemics.
\newblock \emph{Proceedings of the royal society of london. Series A,
  Containing papers of a mathematical and physical character} 115(772):
  700--721.

\bibitem[{Laskar and Kannala(2020)}]{laskar2020data}
Laskar, Z.; and Kannala, J. 2020.
\newblock Data-Efficient Ranking Distillation for Image Retrieval.
\newblock \emph{arXiv preprint arXiv:2007.05299} .

\bibitem[{Liu et~al.(2020)Liu, Gayle, Wilder-Smith, and
  Rockl{\"o}v}]{liu2020reproductive}
Liu, Y.; Gayle, A.~A.; Wilder-Smith, A.; and Rockl{\"o}v, J. 2020.
\newblock The reproductive number of COVID-19 is higher compared to SARS
  coronavirus.
\newblock \emph{Journal of travel medicine} .

\bibitem[{Park et~al.(2019)Park, Kim, Lu, and Cho}]{park2019relational}
Park, W.; Kim, D.; Lu, Y.; and Cho, M. 2019.
\newblock Relational knowledge distillation.
\newblock In \emph{Proceedings of the IEEE Conference on Computer Vision and
  Pattern Recognition}, 3967--3976.

\bibitem[{Ray et~al.(2020)Ray, Wattanachit, Niemi, Kanji, House, Cramer,
  Bracher, Zheng, Yamana, Xiong et~al.}]{ray2020ensemble}
Ray, E.~L.; Wattanachit, N.; Niemi, J.; Kanji, A.~H.; House, K.; Cramer, E.~Y.;
  Bracher, J.; Zheng, A.; Yamana, T.~K.; Xiong, X.; et~al. 2020.
\newblock Ensemble forecasts of coronavirus disease 2019 (covid-19) in the us.
\newblock \emph{MedRXiv} .

\bibitem[{Sanh et~al.(2019)Sanh, Debut, Chaumond, and
  Wolf}]{sanh2019distilbert}
Sanh, V.; Debut, L.; Chaumond, J.; and Wolf, T. 2019.
\newblock DistilBERT, a distilled version of BERT: smaller, faster, cheaper and
  lighter.
\newblock \emph{arXiv preprint arXiv:1910.01108} .

\bibitem[{Thevarajan et~al.(2020)Thevarajan, Nguyen, Koutsakos, Druce, Caly,
  van~de Sandt, Jia, Nicholson, Catton, Cowie et~al.}]{thevarajan2020breadth}
Thevarajan, I.; Nguyen, T.~H.; Koutsakos, M.; Druce, J.; Caly, L.; van~de
  Sandt, C.~E.; Jia, X.; Nicholson, S.; Catton, M.; Cowie, B.; et~al. 2020.
\newblock Breadth of concomitant immune responses prior to patient recovery: a
  case report of non-severe COVID-19.
\newblock \emph{Nature medicine} 26(4): 453--455.

\bibitem[{Tokozume, Ushiku, and Harada(2017)}]{tokozume2017learning}
Tokozume, Y.; Ushiku, Y.; and Harada, T. 2017.
\newblock Learning from between-class examples for deep sound recognition.
\newblock \emph{arXiv preprint arXiv:1711.10282} .

\bibitem[{Tokozume, Ushiku, and Harada(2018)}]{tokozume2018between}
Tokozume, Y.; Ushiku, Y.; and Harada, T. 2018.
\newblock Between-class learning for image classification.
\newblock In \emph{Proceedings of the IEEE Conference on Computer Vision and
  Pattern Recognition}, 5486--5494.

\bibitem[{Wang et~al.(2020)Wang, Li, Wang, and Gong}]{wang2020neural}
Wang, D.; Li, Y.; Wang, L.; and Gong, B. 2020.
\newblock Neural Networks Are More Productive Teachers Than Human Raters:
  Active Mixup for Data-Efficient Knowledge Distillation from a Blackbox Model.
\newblock In \emph{Proceedings of the IEEE/CVF Conference on Computer Vision
  and Pattern Recognition}, 1498--1507.

\bibitem[{Wang, Chen, and Marathe(2019)}]{wang2019defsi}
Wang, L.; Chen, J.; and Marathe, M. 2019.
\newblock DEFSI: Deep learning based epidemic forecasting with synthetic
  information.
\newblock In \emph{Proceedings of the AAAI Conference on Artificial
  Intelligence}, volume~33, 9607--9612.

\bibitem[{Wu, Leung, and Leung(2020)}]{wu2020nowcasting}
Wu, J.~T.; Leung, K.; and Leung, G.~M. 2020.
\newblock Nowcasting and forecasting the potential domestic and international
  spread of the 2019-nCoV outbreak originating in Wuhan, China: a modelling
  study.
\newblock \emph{The Lancet} 395(10225): 689--697.

\bibitem[{Yang et~al.(2017)Yang, Santillana, Brownstein, Gray, Richardson, and
  Kou}]{yang2017using}
Yang, S.; Santillana, M.; Brownstein, J.~S.; Gray, J.; Richardson, S.; and Kou,
  S. 2017.
\newblock Using electronic health records and Internet search information for
  accurate influenza forecasting.
\newblock \emph{BMC infectious diseases} 17(1): 332.

\bibitem[{Yang, Santillana, and Kou(2015)}]{yang2015accurate}
Yang, S.; Santillana, M.; and Kou, S.~C. 2015.
\newblock Accurate estimation of influenza epidemics using Google search data
  via ARGO.
\newblock \emph{Proceedings of the National Academy of Sciences} 112(47):
  14473--14478.

\bibitem[{Yang et~al.(2020)Yang, Zeng, Wang, Wong, Liang, Zanin, Liu, Cao, Gao,
  Mai et~al.}]{yang2020modified}
Yang, Z.; Zeng, Z.; Wang, K.; Wong, S.-S.; Liang, W.; Zanin, M.; Liu, P.; Cao,
  X.; Gao, Z.; Mai, Z.; et~al. 2020.
\newblock Modified SEIR and AI prediction of the epidemics trend of COVID-19 in
  China under public health interventions.
\newblock \emph{Journal of Thoracic Disease} 12(3): 165.

\bibitem[{Zhang et~al.(2017)Zhang, Cisse, Dauphin, and
  Lopez-Paz}]{zhang2017mixup}
Zhang, H.; Cisse, M.; Dauphin, Y.~N.; and Lopez-Paz, D. 2017.
\newblock mixup: Beyond empirical risk minimization.
\newblock \emph{arXiv preprint arXiv:1710.09412} .

\end{thebibliography}
\bibstyle{aaai21}

\end{document}